\def\year{2018}\relax
\documentclass[letterpaper]{article} %DO NOT CHANGE THIS
\usepackage{aaai18}  %Required
\usepackage{times}  %Required
\usepackage{helvet}  %Required
\usepackage{courier}  %Required
\usepackage{url}  %Required
\usepackage{graphicx}  %Required
\usepackage{amsmath}
\usepackage{makecell}

\newcommand{\newcite}[1]{\citeauthor{#1} (\citeyear{#1})}

\newcommand{\vect}[1]{\mathbf{#1}}
\newcommand{\matr}[1]{\mathbf{#1}}

\newcommand{\vc}[0]{\vect{c}}
\newcommand{\ve}[0]{\vect{e}}

\newcommand{\vh}[0]{\vect{h}}

\newcommand{\vx}[0]{\vect{x}}
\newcommand{\vz}[0]{\vect{z}}

\newcommand{\vs}[0]{\vect{s}}

\newcommand{\vy}[0]{\vect{y}}

\newcommand{\vr}[0]{\vect{r}}

\newcommand{\mW}[0]{\matr{W}}

\newcommand{\mE}[0]{\matr{E}}

\newcommand{\mU}[0]{\matr{U}}

\newcommand{\mS}{\matr{S}}

\begin{document}

\title{Style Transfer in Text: Exploration and Evaluation}
\author{
Zhenxin Fu\textsuperscript{1}, Xiaoye Tan\textsuperscript{1}, Nanyun Peng\textsuperscript{2},
 Dongyan Zhao\textsuperscript{1,3}, Rui Yan\textsuperscript{1,3}\thanks{Corresponding author: Rui Yan (ruiyan@pku.edu.cn)} \\
\textsuperscript{1}{Institute of Computer Science and Technology, Peking University, Beijing, China}\\
\textsuperscript{2}{Information Science Institute, University of Southern California, California, USA}\\
\textsuperscript{3}{Beijing Institute of Big Data Research, Beijing, China} \\
\{fuzhenxin, txye, zhaodongyan, ruiyan\}@pku.edu.cn, npeng@isi.edu
}

% abstract
\maketitle
\begin{abstract}
%Style transfer is an important problem in natural language processing (NLP). 
The ability to transfer styles of texts or images, is an important measurement of the advancement of artificial intelligence (AI).
However,  the progress in language style transfer is lagged behind other domains, 
such as computer vision, mainly because of the lack of parallel data and reliable evaluation metrics. 
In response to the challenge of lacking parallel data, we explore learning style transfer from non-parallel data. 
We propose two models to achieve this goal. 
The key idea behind the proposed 
models is to learn separate content representations and style representations using adversarial networks. 
Considering the problem of lacking principle evaluation metrics, we propose two novel evaluation metrics that measure two aspects of style transfer: 
transfer strength and content preservation. 
We benchmark our models and the evaluation metrics on two style transfer tasks: paper-news title transfer, 
and positive-negative review transfer.  
Results show that the proposed content preservation metric is highly correlate to human judgments,  
and the proposed models are able to generate sentences with similar content preservation score but higher style transfer strength comparing to auto-encoder.
\end{abstract}
\noindent 

% Introduction
%\vspace{-2em}
\section{Introduction}
Style transfer is an important problem in many subfields of artificial intelligence (AI), such as natural language processing (NLP) and computer vision \cite{gatys2016image,gatys2016preserving,zhu2017unpaired,li2017demystifying}, as it reflects the ability of intelligence systems to generate novel contents.
Specifically, style transfer of natural language texts is an important component of natural language generation. It facilitates many NLP applications, such as automatic conversion of paper title to news title, which reduces the human effort in academic news report. % Chanllenges
For tasks like poetry generation \cite{yan2013poet,yan2016poet,ghazvininejad2016generating}, style transfer can be applied to generate poetry in different styles.
Nevertheless, the progress in style transfer of language is lagged 
behind other domains such as computer vision, largely because of the lack 
of parallel corpus and reliable evaluation metrics. 

%importance of parallel data
Sequence to sequence (seq2seq) neural network models \cite{sutskever2014sequence} have demonstrated great success in many generation tasks, 
such as machine translation, dialog system and image caption, with the requirement of a large amount of parallel data. 
However, it is hard to get parallel data for tasks such as style transfer. 
For instance, there is only a small number of academic news reports which have corresponding papers.
Therefore, we need algorithms that perform style transfer without parallel data. 

Another major challenge of style transfer is to separate style from the content. In computer vision, \newcite{li2017demystifying} proposes an expression to distinguish style and content of a picture. However, this is under-explored in the NLP community.
How to separate content from style in text remains an open research problem in text style transfer. 

%Evaluation
Evaluation is also a key challenge in style transfer. In machine translation and summarization, researchers use BLEU \cite{papineni2002bleu} and ROUGE \cite{lin2004rouge} to compute the similarity between model outputs and the ground truth. However, we lack parallel data for style transfer to provide ground truth references for evaluation. The same problem also exists in style transfer in computer vision. 
To solve this problem, we propose a general evaluation metric for style transfer in natural language processing. There are two aspects of the evaluation metric; one is {\it transfer strength} and the other is {\it content preservation}.

%model introduction
In this paper, we explore two models for text style transfer, to approach the aforementioned problems of 1) lacking parallel training data and 2) hard to separate  the style from the content. 
The models achieve the goals by multi-task learning \cite{caruana1998multitask} and adversarial training \cite{goodfellow2014generative} of deep neural network.
The first model implements a multi-decoder seq2seq proposed by \newcite{sutskever2014sequence}, 
where the encoder is used to capture the content $c$ of the input $X$, and the 
multi-decoder contains $n ( n \geq 2)$ decoders to generate outputs in different styles. 
The second model uses the same encoding strategy, but introduces style embeddings that are jointly 
trained with the model. The style embeddings are used to augment the encoded representations, so 
that only one decoder needs to be learned to generate outputs in different styles. 

%Results
The experiments on two tasks: paper-news title transfer and positive-negative review transfer showed that 
each of the proposed model has its own strength and can be used in different transfer requests, and the proposed content preservation metric has a high correlation with human judgments.

%Contributions
\section{Contributions}
The contributions of this paper are three fold:
\begin{itemize}
\item We compose a dataset\footnote{Available at https://github.com/fuzhenxin/textstyletransferdata} of paper-news titles to facilitate the research in language style transfer. 
\item We propose two general evaluation metrics for style transfer, 
which considers both transfer strength and content preservation. 
The evaluation metric is highly correlated to the human evaluation. 
\item We proposed and evaluated two models for learning style transfer without parallel corpora.  
The proposed models addressed the key challenge of lacking parallel data for training in style transfer,  
and each model has its own advantages under different scenarios.

\end{itemize}

%figure_model
\begin{figure*}[htb]
\centering
\includegraphics[width = .85\textwidth]{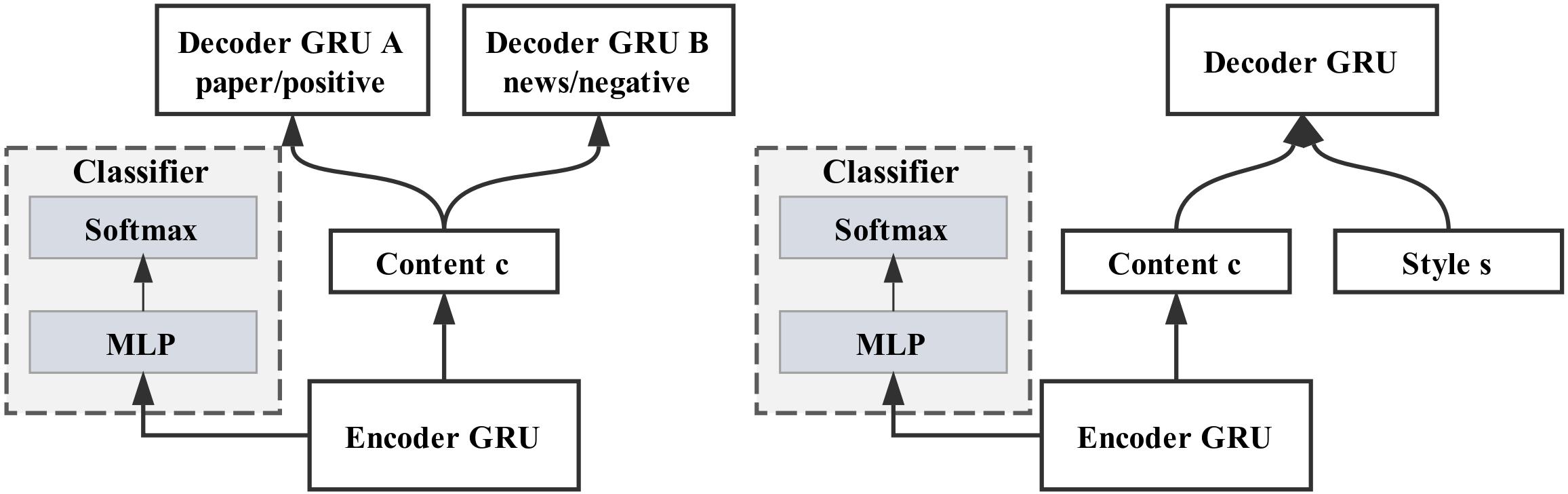}
\caption{Two models in this paper, multi-decoder (left) and style-embedding (right). 
   Content $c$ represents output of the encoder. 
   Multi-layer Perceptron (MLP) and Softmax constitute the classifier. 
   This classifier aims at distinguishing the style of input $X$. 
   An adversarial network is used to make sure content $c$ does not have style representation. 
   In style-embedding, content $c$ and style embedding $s$ are concatenated and $[c,e]$ is fed into decoder GRU. }
\label{figure_model}
\end{figure*}

%Related Work
\noindent
\section{Related Work}

%Style Transfer in Computer Vision
\subsection{Style Transfer in Computer Vision}
In recent years, style transfer has made significant progress in computer vision. 
\newcite{gatys2016image} separated the content and style of images and recombined them to generate new images. 
\newcite{gatys2016preserving} designed a simple linear model to change the color of the pictures.
Their methods use only one image to represent a style.
However, it does not work in NLP because a single sentence or a short article does not store enough style information.

\newcite{zhu2017unpaired} proposes CycleGAN to do image-image translation. 
It firstly learns a mapping $G: X \to Y$ using an adversarial loss, 
and then a reverse mapping $ F: Y \to X$ with a cycle loss $F(G(X)) \approx X$ which performs unpaired image to image translation. 
CycleGAN shows qualitative results, nevertheless, discrete text is hard to implement cycle training. 
\newcite{li2017demystifying} proposes  
to treat style transfer as a domain adaptation problem. 
They theoretically show that Gram metrics is equivalent to minimize the Maximum Mean Discrepancy (MMD) for image. 
But there is no evidence showing similar metric works on text.

%Style Transfer in Natural Language Processing
\subsection{Style Transfer in Natural Language Processing}
\newcite{jhamtani2017shakespearizing} explores automatic methods to transform text from modern English to 
Shakespearean English using parallel data. The model was based on seq2seq and enriched it with pointer network \cite{vinyals2015pointer}.
They used a modern-Shakespeare word dictionary to form candidate words for pointer network, 
however, paired-word dictionary is a scares resource that does not exist in most style transfer tasks,
and it required parallel corpora.

There are previous work on style transfer without parallel data. 
\newcite{mueller2017sequence} proposed a variational auto-encoder (VAE) based model 
to revise a new sequence to improve its associated outcome.
However, there is no significative evaluation for style transfer.
It uses non-parallel data.
\newcite{shen2017style} explored style transfer for sentiment modification,
decipherment of word substitution ciphers and recovery of word order. 
They used VAE as the base model and used an adversarial network to align different styles. 
However, their evaluation only considered the classification accuracy.  
We argue that content preservation is another indispensable evaluation metric for style transfer. 

Other threads of work that are closely related to us including style analysis and style-controlled text generation.
\newcite{braud2017writing} explores many types of features for style prediction, ranging from n-grams to discourse,  
and found that simple models performed well. 
\newcite{ficler2017controlling} controls linguistic style of generated text 
using conditioned recurrent neural networks (CRNN). The major difference between these work and ours is 
that they do not have source sentences where we need to transfer the style.

%Adversarial Network in domain separation
\subsection{Adversarial Networks for Domain Separation}
Adversarial networks have been successfully applied to domain separation problems. 
\cite{ganin2015unsupervised} proposed deep domain adaptation approach to encourage domain-invariant features.
This model can be trained on labeled source domain data and unlabeled target domain data. 
\cite{bousmalis2016domain} used  adversarial networks to learned shared representations 
between two domains which don't contain the individual features of each domain. 
\cite{chen2017adversarial} proposed a multi-task framework to generate shared and private representations for sentences.  
The shared layer is also reinforced by adversarial networks. 
\cite{long2016deep} proposed a joint adaptation network, which adopted the adversarial strategy 
to maximize
joint maximum mean discrepancy. 
The major difference between these work and ours is that they do not need to generate 
new sentences. How adversarial networks work on controlled generation is largely untested.

%Model
\section{Model}
We propose two models for style transfer in this paper: multi-decoder and style-embedding. 
Both models are based on the neural sequence to sequence model. 
The common ground  of the two models is to learn a representation for the input 
sentence that only contains the content information. 
Then the multi-decoder model uses different decoders, one for each style, 
to generate texts in the corresponding style. 
The style-embedding model, in contrast, learns style embeddings addition 
to the content representations. Then a single decoder is trained to generate 
texts in different styles based on both the content representation and the style embedding.
Figure \ref{figure_model} illustrates the two models. We give more details about each model in the following sections.

%auto-encoder
\subsection{Background: Auto-encoder Seq2seq Model}
Auto-encoder \cite{rumelhart1985learning} is a type of neural networks that 
learns a hidden representation for the input. It was mainly used for dimension reduction 
in the past, but more recently, the concepts have been widely used for generative models. 
In the auto-encoder seq2seq model, an encoder is learned to generate intermediate representation 
of input sequence $X=(x_1, \dots,x_{T_x})$ of length $T_x$. Then a decoder is trained to 
recover the input $X$ using the intermediate representation. 
For the style transfer problem, we use the auto-encoder seq2seq model as our base model, since we 
expect minimum changes from the input to the output. We give more details about this model 
as we also use the components of this model in our proposed models.

%\vspace{-1em}
\paragraph{Encoder} In auto-encoder seq2seq model, both the encoder and decoder are recurrent neural networks (RNNs).
We employ the gated recurrent unit (GRU) variant which uses gates to control the information flow.
A GRU unit is composed of the following components:
{\small
\begin{align}
     \vs_j =& ~ \vz_j \odot \vh_j + (1-\vz_j) \odot \vs_{j-1}, \\
     \vh_j =& ~\text{tanh} \left(   \mW \mE[x_{j-1}] + \vr_j \odot (\mU\vs_{j-1})  \right), \\
	\vr_j =& ~ \sigma \left(  \mW_r \mE[x_{j-1}] + \mU_r \vs_{j-1}  \right), \\
	\vz_j =& ~ \sigma \left(  \mW_z \mE[x_{j-1}] + \mU_z \vs_{j-1}  \right), 
\end{align}
}%
where $\vs_j$ is the activation of GRU at time $j$; $\vh_j$ is an intermediate state computes the candidate activation. 
$r_j$ is a reset gate that controls how much to reset from the previous activation for the candidate activation. 
Similarly, $z_j$ is an update gate that controls how much to update the current activation based on the previous 
activation and the candidate activation. 
$\mE$ is a word embedding matrix that is used to convert the input words to vector representations.  
$\mE$, $\mW$, $\mU$, $\mW_r$, $\mU_r$, $\mW_z$, $\mU_z$ are model parameters. We use $\Theta_e$ to denote 
all the parameters of the encoder,  then the encoder can be abstracted as:
\begin{equation}
\small
\mS = Encoder(\vx; \Theta_e)
\end{equation}
%\vspace{-2em}
\paragraph{Decoder} The decoder takes the last state of the encoder to start the generation process. 
It generates tokens by predicting the most probable next token based on previous tokens.
The probability of an output sequence given an input $P(\vy_i|\vx_i)$ is defined by Equation~\ref{possibilityauto-encoder}, 
where $i$ indexes the instances, $j$ the output tokens. 
The probability $p(.)$ of generating each token can be computed by the softmax function.
{\small
\begin{align}
	&P \left(\vy_i|\vx_i; \Theta_d \right) =            \notag \\
     &~ \prod_{j=1}^{T_y}p\left ( y_{i,j}|Encoder(\vx_i; \Theta_e),y_{i,1},\dots,y_{i,j-1};\Theta_d \right )          \label{possibilityauto-encoder}
\end{align}
}%
The loss function of the encoder-decoder seq2seq model (Equation~\ref{lossauto-encoder}) minimizes the negative log probability of the training data, 
where $M$ denotes the size of the training data, 
$\Theta_e$ and $\Theta_d$ are the parameters of the encoder and the decoder, respectively. The model can be trained end-to-end.
{\small
\begin{align}
	L_{seq2seq} \left( \Theta_e,\Theta_d \right) =& -\sum_{i=1}^{M}\log~P\left(\vy_i|\vx_i;\Theta_e,\Theta_d \right)  \label{lossauto-encoder}
\end{align}
}%
In auto-encoder, we let the output sequence $\vy$ to be the same as the input sequence $\vx$.

%Multi-decoder
\subsection{Multi-decoder Model}
The multi-decoder model for style transfer is similar to an auto-encoder with several decoders, with the 
exception that the encoder now tries to learn some content representations that do not reflect styles. 
The style specific decoders (one for each style) then take the content representations and generate 
texts in different styles. The challenge of this model is how to generate content representation $\vc$ from input $\vx$. 
In the original auto-encoder model, the encoder generates representations that contain both content and style information.

\newcite{chen2017adversarial} used an adversarial network to separate the shared and the private features for multi-task learning to help chinese word segmentation. 
We use a similar adversarial network to separate the content representation $\vc$ from the style.
The adversarial network is composed of two parts.
The first part aims at classifying the style of $\vx$ given the representation learned by the encoder.
The loss function minimizes the negative log probability of the style labels in the training data, as denoted in Equation~\ref{lossclassifier}: 
{\small
\begin{align}
&L_{adv1}\left(\Theta_c \right)= - \sum_{i=1}^{M}\log~p \left( l_i|Encoder(\vx_i; \Theta_e);\Theta_c \right),    \label{lossclassifier} 
\end{align}
} %
where $\Theta_c$ is the parameters of a multi-layer perceptron (MLP) for predicting the style labels. 
The second part of the adversarial network aims at making the classifier unable to identify the style of $\vx$ by 
maximize the entropy (minimize the negative entropy) of the predicted style labels, as denoted in Equation~\ref{lossadv2}. 
{\small
\begin{align}
&L_{adv2}\left( \Theta_e \right)=-\sum_{i=1}^{M}\sum_{j=1}^{N} H\left( p \left( j|Encoder(\vx_i; \Theta_e);\Theta_c \right) \right), \label{lossadv2}  
\end{align}
} %
where $\Theta_e$ is the parameters of the encoder and $N$ is the number of styles, as introduced in previous sections. 
Note that the two parts of the adversarial network update different sets of parameters, and
they work together to make sure that outputs of encoder $Encoder(\vx_i; \Theta_e)$ do not contain style information. 

%Introduction to loss function
While the encoder is trained to produce content representations, the multiple decoders are trained to 
take the representations produced by the encoder and generate outputs in different styles. 
The loss function for each decoder is similar to  Equation~\ref{lossauto-encoder}, and the total generation 
loss is the sum of the generation loss of each decoder, as defined in Equation~\ref{seq2seq1}.
{\small
\begin{align}
&L_{gen1} \left( \Theta_e,\Theta_d \right) = \sum_{i=1}^{L} L_{seq2seq}^i \left( \Theta_e,\Theta_d^i \right) \label{seq2seq1}
\end{align}
}%
The final loss function of the multi-decoder model is composed of three parts: 
two for the adversarial network and one for the sequence to sequence generation.   
It simply takes an unweighted sum of the three parts as illustrated in Equation~\ref{totalloss1}.
{\small
\begin{align}
&L_{total1}\left( \Theta_e,\Theta_d,\Theta_c \right)  \notag \\
&= L_{gen1}\left( \Theta_e,\Theta_d \right) + L_{adv1}\left( \Theta_c \right) + L_{adv2}\left( \Theta_e \right)  \label{totalloss1}
\end{align}
}%

%Style-embedding
\subsection{Style-embedding Model} 
Our second model uses style embeddings to control the generated styles. 
This is inspired by \cite{li2016persona}, which proposed a model to embed 
personal information into vector representations for persona-conversation,
and \cite{ficler2017controlling} which generated text with different contents and styles 
using conditional RNNs that conditioned on both content and style parameters.

In this model, the encoder and the adversarial network parts are the same as the  
multi-decoder model, to generate content representations $\vc$.
In addition, style embeddings $\mE \in R^{N\times d_s}$ are introduced to represent the styles, 
where $N$ denotes the number of styles and $d_s$ is the dimension of style embedding.
A single decoder is trained in this model, which takes the concatenation of the content representation 
$\vc$ and the style embedding $\ve$ of a sentence as the input to generate texts in different styles.

The loss function of the style-embedding model is defined in (\ref{totalloss2}), 
where $L_{gen2}$ is the loss function for the seq2seq generation very similar to Equation~\ref{lossauto-encoder}. 
The only difference is that it also contains the parameter $\mE$ for style embeddings, that are 
jointly trained with the mode.
The total loss is similar to the multi-decoder model in Equation~\ref{totalloss1},  
where $L_{adv1}$ and $L_{adv2}$ are the same as in Equations~\ref{lossclassifier} and ~\ref{lossadv2}. 
{\small
\begin{align}
&L_{total2}\left( \Theta_e,\Theta_d,\Theta_c,\mE \right)  \notag \\
&= L_{gen2}\left( \Theta_e,\Theta_d,\mE \right) + L_{adv1}\left( \Theta_c \right) + L_{adv2}\left( \Theta_e \right)  \label{totalloss2}
\end{align}
}%

%\subsection{Discussion}
%Both the two models proposed in this section can be viewed as a multi-task extension to the auto-encoder 
%seq2seq model. 
%In the multi-decoder model, the multi-task learning is achieved by having the different tasks 
%sharing the same encoder and have their own decoders. 
%In the style-embedding model, on the other hand, the multi-task learning is achieved by introducing 
%different style embeddings for different tasks. The rest parts including the encoder and the decoder 
%are all shared among the tasks.
%Since the style-embedding model shares more parameters among tasks, less training data is needed to 
%train the model, but the style embeddings have heavier burden to encode the style information.

\subsection{Parameter Estimation}
We use Adadelta \cite{zeiler2012adadelta} with the initial learning rate 0.0001 and batch size 128 to learn the parameters for all models. 
The best parameters are decided based on the perplexity on the validation data with a maximum of 50 training epochs for paper-news task and 10 training epochs for positive-negative task. 

For the multi-decoder model, we train the multiple decoders alternately, using the data 
in the corresponding style. For the style-embedding model, we randomly shuffled the data during training, 
and jointly learned the style embeddings with the encoder-decoder part.

\section{Evaluation}
Evaluation plays an important role in style transfer. 
Automatic evaluation metrics speed up development. 
And they provide criteria to compare different models. 

BLEU \cite{papineni2002bleu} is a popular evaluation metric in neural machine translation 
and ROUGE \cite{lin2004rouge} is popular in text summarization. 
NIST \cite{doddington2000nist} and Meteor \cite{banerjee2005meteor} are also used widely in Natural Language Processing. 
They evaluate the similarity between model output and ground truth by word overlapping. 
AM-FM \cite{banchs2011fm} proposes an automatic evaluation for NMT without ground truth. 
This model computes sentence embedding first and then computes cosine similarity between source language input and target language output.
It gets sentence embedding by Singular Value Decomposition (SVD), which trains source and target language together.
RUBER \cite{tao2017ruber} was proposed to evaluate dialog system, it divides evaluation into referenced and unreferenced part. 
In referenced part, it calculates the similarity between model output and ground truth by cosine distance of sentence embedding.

We propose two general evaluation metrics, one is transfer strength, 
the other one is content preservation.

%Transfer Strength
\subsection{Transfer Strength}
The main task of this model is to transfer source style to target style, 
so transfer strength evaluates whether the style is transferred. 
We define the metric as transfer strength and implement it using a classifier. 
There are more than 100,000 training data for this task. 
We use a LSTM-sigmoid classifier which performs well in big data. 
The style is defined in (\ref{lstmoutput}). 
This classifier is based on keras examples\footnote{https://github.com/fchollet/keras/blob/\\master/examples/imdb\_lstm.py}. 
Transfer strength accuracy is defined as $\frac{N_{right}}{N_{total}}$, ${N_{total}}$ is the number of test data, 
and ${N_{right}}$ is the number of correct case which is transferred to target style.
{\small
\begin{align}
l_{style}=
\begin{cases} paper(positive) &output \leq 0.5   \label{lstmoutput} \\
news(negative)  &output>0.5
\end{cases}
\end{align}
}%

For similar task, \cite{shen2017style} uses classifier to evaluate style transfer. \cite{zhou2017emotional} controls emotion of conversation, it also uses a classifier to evaluate chatbot generated emotional response.

%Content preservation
\subsection{Content Preservation}
Another important aspect of style transfer is content preservation. 
It is easy to train a model that has 100\% transfer strength by only generating the target style words. 
Therefore, we propose a metric for content preservation, which can evaluate the similarity between source text and target text. 
Content preservation rate is defined as cosine distance (\ref{cosinedistance}) between source sentence embedding $v_s$ 
and target sentence embedding $v_t$. 
Sentence embedding consists of max,min,mean pooling of word embedding defined in (\ref{poolingconcate}).
{\small
\begin{align}
v_{min}[i]=&~ \min\{w_1[i], \dots w_n[i]\} \\
v_{mean}[i]=&~ \text{mean}\{w_1[i], \dots w_n[i]\} \\
v_{max}[i]=&~ \max\{w_1[i], \dots w_n[i]\} \\
v =&~ [v_{min}, v_{mean}, v_{max}]                                                    \label{poolingconcate}     \\
score=&~ \frac{v_s^{\top} v_t}{\left \| v_s \right \| \cdot \left \| v_t \right \|}   \label{cosinedistance}     \\
score_{total} = &~ \sum_{i=1}^{M_{test}}score_{i}
\end{align}
}%
For word embedding, we use pre-trained Glove \cite{pennington2014glove} published at 
stanford nlp\footnote{https://nlp.stanford.edu/projects/glove/}. 
This project contains word embedding trained on 6 billion tokens, 
containing 400k vocabularies, with dimension 50, 100, 200 and 300. 
In our model, we use dimension 100.

Although a single integrated metric that combines transfer strength and content preservation as F1 score 
seems plausible to measure the performance of the systems, it is not the best for style transfer, since sometimes the transfer 
strength is more important, while in other cases the content preservation is the focus.
A weighted integration would be ideal for different scenarios. We leave the weighted integration for the 
future work and report both metrics in this paper.

\begin{table}[t]
\centering
\resizebox{.45\textwidth}{!}{  
\begin{tabular}{|c|c|c|c|c|}
\hline
dataset             & \multicolumn{2}{c|}{title}  & \multicolumn{2}{c|}{review} \\ \hline
style type          & paper        & news         & positive      & negative     \\ \hline
\#sentences                & 107,538      & 108,503      & 400,000      & 400,000      \\ \hline
vocabulary size     & \multicolumn{2}{c|}{80,000} & \multicolumn{2}{c|}{60,000}  \\ \hline
\end{tabular}
}
\caption{Size of datasets}
%\vspace{-1.5em}
\label{table_data}
\end{table}

%Experiments
\section{Experimental Setup}
%Datesets

\subsection{Datasets}
We used two datasets to evaluate the performances of the proposed methods. One is the paper-news title dataset, 
the other is the positive-negative review dataset; both are non-parallel corpora. 
We composed the first dataset ourselves and used the data released by \newcite{he2016ups} 
as the second dataset. 
For both datasets, we divided them into three parts: training, validation, and test data. 
The size of the validation and test data is 2,000 sentences, and the rest are used as training data.
And the partition is the same between model and evaluation.
 
We ignored the sentences that contain more than 20 words, 
and converted all characters to lower cases. We also replace all the numbers to a special 
string ``$\langle $NUM$ \rangle$" as a preprocessing step.
Some statistics about the datasets is summarized  in Table \ref{table_data}. 

%Paper-News Title
\subsubsection{Paper-News Title Dataset}
In this dataset, the paper titles are crawled from academic websites including ACM Digital Library\footnote{http://dl.acm.org}, 
Arxiv\footnote{https://arxiv.org}, Springer\footnote{https://link.springer.com}, 
ScienceDirect\footnote{http://www.sciencedirect.com}, and Nature\footnote{https://www.nature.com}. 
The news titles are from UC Irvine Machine Learning Repository \cite{Lichman:2013}, which contains 
422,937 news titles. We filtered it down to 108,503 titles which belong to science and technology category. 

%positive-negative Review
\subsubsection{Positive-Negative Review Dataset}
This dataset contains Amazon product reviews published by \newcite{he2016ups}. 
It contains 142,800,000 product reviews from 1996 to 2014 in Amazon, which span the domains of
books, electronics, movies, etc. 
We randomly select 400,000 positive and 400,000 negative reviews to compose our dataset.

%figure_result
\begin{figure*}[htb]
\centering
\includegraphics[width = .5\textwidth]{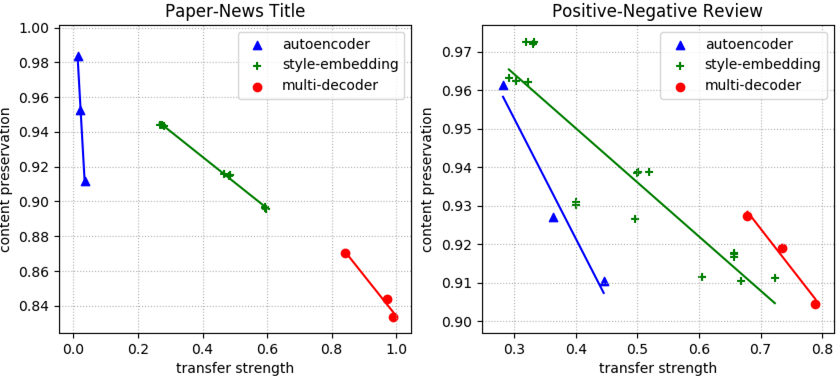}

\caption{Results for auto-encoder, multi-decoder and style embedding for two tasks, 
paper-news title style transfer (left) and positive-negative review style transfer (right).
Different nodes for the same model denote different hyper-parameters.}
\label{figure_result}
\end{figure*}

%figure_correlation
\begin{figure}[]
\includegraphics[width = .5\textwidth]{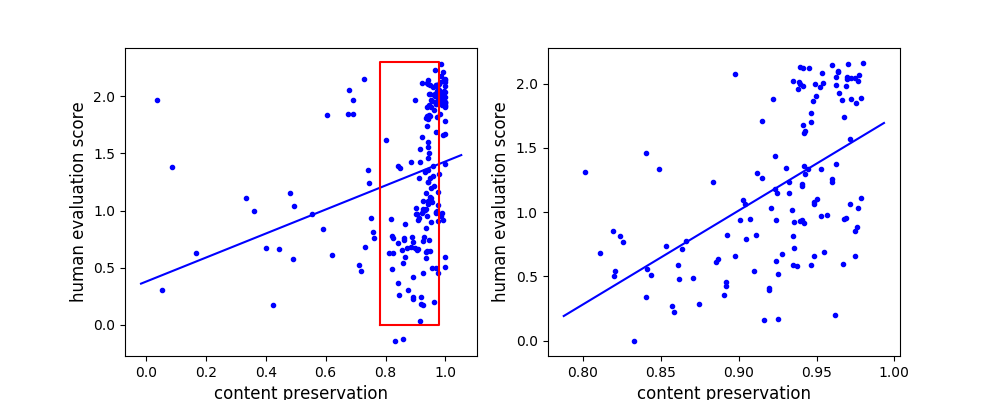}
\caption{Score correlation of content preservation and human evaluation. Gaussian noise is added to human evaluation for better visualization. The partial enlarged graph is shown on the right.}
\label{figure_correlation}
\end{figure}

\subsection{Model Settings} 

Since this is an exploratory paper, we compare several parameters settings instead 
of trying to find a single set of ``best parameters".
For paper-news title transfer, we explored word embedding size of 64, 
encoder hidden vector size among \{32,64,128\}, and style embedding size among \{32,64,128\}.
For positive-negative review transfer, we explored word embedding size 
of 64 for multi-decoder and \{64,128\} for style-embedding model, encoder hidden vector size among \{16,32,64\},
and style embedding size among \{16,32,64\}. 

\subsection{Evaluation Settings}
As is introduced in previous sections, an LSTM-sigmoid classifier is needed to measure the transfer strength. 
We train an LSTM with the input word embedding dimension and hidden state dimension both be 128. 
On the paper-news title transfer dataset, the training stops after two epochs, and the accuracy on the validation data is 98.8\%.
For the positive-negative review dataset, the training also stops after two epochs, with an accuracy of 84.8\% on validation.

For the content preservation metric, we use pretrained 100-dimensional word embeddings to compute sentence similarities. 
For the positive and negative review transfer task, we filter out the sentiment words to make sure the content preservation metric  
indeed measures the content similarity. 
A positive and negative word dictionary is used to conduct the filtering. 

%Results and Analysis
\section{Results and Analysis}

As we discussed in previous sections, this paper is exploratory. We are exploring whether we can learn style transfer with 
non-parallel data, and whether we can define some evaluation metrics to measure how well the models do in text style transfer.
Therefore, we first examine how does the proposed evaluation metrics compare to human judgments.

%Human evaluation
\subsection{Comparison with Human Judgments} 
To ensure our proposed content preservation metric is efficient in measuring the sentence similarities, we compare it against human judgments. 
The human judgments are obtained by randomly sampling 200 paper-news transferred pair from the test data,
target transferred sentences are generated by style-embedding model, 
and ask three different people to rate the pairs with scores \{0, 1, 2\}. 
2 means the two sentences are very similar; 1 means the two sentences 
are somewhat similar; and 0 indicates the two sentences are not similar. 
We conduct this experiment on Amazon Mechanical Turk\footnote{https://www.mturk.com/}. 
The scores for each pair from different people are averaged to generate the final human judgment scores. 
We then calculate the Spearman's coefficient (a measurement for accessing monotonic relationships) between the human judgment scores 
and our content preservation metric. 
The correlation score is 0.5656 with p-value$<$0.0001, indicates a high correlation between human judgment scores and the 
content preservation metric.  
Figure \ref{figure_correlation} illustrates the correlation.

%table_show_case
\begin{table*}[t]
\centering
\begin{tabular}{l}
\hline
source~~ positive: all came well sharpened and ready to go .                                                             \\
auto-encoder:~~~~~~  $\to$negative: all came well sharpened and ready to go .            \\
multi-decoder:~~~~~  $\to$negative: all came around , they did not work .            \\ 
style-embedding:~  $\to$negative: my $\langle $NUM$ \rangle$ and still never cut down it .       \\ \hline
source~~ negative: my husband said it was obvious so i had to return it .                                                \\
auto-encoder:~~~~~~~  $\to$positive: my husband said it was obvious so i had to return it .              \\
multi-decoder:~~~~~  $\to$positive: my husband was no problems with this because i had to use .       \\ 
style-embedding:~  $\to$positive: my husband said it was not damaged from i would pass right .         \\ \hline
source~~ paper: an efficient and integrated algorithm for video enhancement in challenging lighting conditions          \\ 
auto-encoder:~~~~~~~  $\to$news: an efficient and integrated algorithm for video enhancement in challenging lighting conditions\\
multi-decoder:~~~~~  $\to$news: an efficient and integrated and google smartphone for conflict roku together wrong   \\
style-embedding:~  $\to$news: an efficient and integrated algorithm, for video enhancement in challenging power worldwide \\ \hline
source~~ news: luxury fashion takes on fitness technology                                                      \\
auto-encoder:~~~~~~~  $\to$paper: luxury fashion takes on fitness technology                            \\
multi-decoder:~~~~~ $\to$paper: foreign banking carbon on fitness technology              \\
style-embedding:~  $\to$paper: luxury fashion algorithms on fitness technology          \\ \hline
\end{tabular}
\caption{Case study of style transfer}
\label{table_show_case}
\end{table*}

%Paper-News Task
\subsection{Model Performances}
We then explore the effect of different parameters on different models for style transfer. 
Figure~\ref{figure_result} gives an overview of the results. 
We can see that in both tasks and all the models, transfer strength and content preservation are 
negatively correlated. This indicates that within the same model, to get more style changes, 
one has to lose some contents. We also see the slopes of the trade-off curves appear to be 
less steep in our proposed models than in the auto-encoder, which indicates our 
models strike a better balance between the two aspects (transfer strength and content preservation) of style transfer. 
We now give detailed analysis  of the performances of different models with different parameters on the two tasks, respectively. 
More details about the influences of the hyper-parameters can be found at \url{https://arxiv.org/abs/1711.06861}. 

\paragraph{Paper-News Title Transfer} For the paper-news title transfer task, the auto-encoder is able to recover most of the content,  
but with few transfer strength, just as we expected. 
The multi-decoder performs better on transfer strength, while style-embedding performs better on content preservation. 
Both are also able to achieve considerably high scores in two metrics, so there is no clear winning model.

More specifically, for the style-embedding model, the transfer strength ranges from 0.2 to 0.6 when using different hyper-parameters, 
and the content preservation ranges from 0.89 to 0.95.  Both cover a wide range and would be useful for certain downstream tasks.
For the multi-decoder model, it generally tends to generate results with high transfer strength but low content preservation. 
Therefore, we suggest using multi-decoder and style-embedding  in different request scenarios.

%Positive-Negative Task
\paragraph{Positive-Negative Review Transfer}
For the positive-negative review transfer task, the transfer strength of auto-encoder is no longer nearly zero like in the paper-news title transfer task, 
probably because the classifier used to measure the transfer strength is not perfect\footnote{The accuracy of this classifier is only 84.8\% 
on the validation data, probably because some sentences in the positive-negative review dataset do not have significant sentiment}. 
The transfer strength measure is not as reliable as it is in the paper-news title task. 

For the style-embedding model, it covers a quite wide range in both transfer strength and content preservation. 
The multi-decoder model still shows high transfer strength as is in the paper-news title transfer task, and it achieved higher content preservation than 
that in paper-news title transfer. 
In this dataset, the multi-decoder model performs better than the style-embedding model on both metrics (the red line is on the upper right over the green line).  

%Analysis in multi-task view
\subsection{Analysis in a Multi-task Learning View}
Auto-encoder, style-embedding and multi-decoder can be seen as different strength implement of multi-task learning. In our model, generating different titles can be seen as different tasks. For some kind of multi-task learning, different tasks share parameters to share features in different tasks.

For auto-encoder, two tasks share all the parameters, so it does not have the ability to generate different style sequence. For style-embedding, two tasks share encoder and decoder with separate style embedding, so it has weak ability to generate different style sequence. For multi-decoder, two tasks share encoder with two separate decoders, so it shows high ability to generate different style sequence. For content preservation, more parameters are shared, less distinction between two tasks and more content is preserved.
Since the style-embedding model shares more parameters among tasks, less training data is needed to train the model, but the style embeddings have heavier burden to encode the style information.

%Lower Bound of Content Reservation
\subsection{Lower Bound for Content Preservation}
We also estimate the lower bound of the content preservation metric, to gauge how well our model performed in preserving the content. 
The lower bound is estimated by randomly sampling 2,000 sentence pairs from the two datasets, respectively. 
Results show that the estimated lower bound of content preservation on the paper-news title dataset is 0.609 and 0.863 on the positive-negative review dataset. 
For both datasets, our models achieved much higher content preservation scores than the lower bound. 
This indicates that the proposed model learned to preserve the content of the source sentence well.

\subsection{Qualitative Study}
To give people some intuitive sense about how our models perform, we  sampled one instance from each style transfer case,  
and show the results of three models in Table \ref{table_show_case}. 
We can see that the auto-encoder almost always produce the identical output text as the input. 
The other two models tend to generate results that replace a few significant words or phrases,  
but preserve most of the content. 
Both models perform quite well on the positive-negative style transfer, but less well on the paper-news transfer.

%Conclusions
\section{Conclusions}
We studied the problem of style transfer with non-parallel corpora. We proposed two models and 
two evaluation metrics to advance the research in this area. We also composed two datasets: paper-news title dataset  
and positive-negative review dataset, to gauge the efficiency of the proposed models and evaluation metrics. 
Experiments showed that the proposed models can be used to learn style transfer from non-parallel data, and the proposed 
content preservation evaluation metric is highly correlated to human judgment. 

In the future, we plan to propose more comprehensive evaluation metrics (including sentence fluency) and conduct through study with human evaluation, to better shape the research in style transfer.

\section{Acknowledgment}
We thank Jin-ge Yao for discussions on this paper. This work was supported by the National Key Research and Development Program of China (No. 2017YFC0804001), the National Science Foundation of China (No. 71672058), and Contract W911NF-15-1-0543 with the US Defense Advanced Research Projects Agency (DARPA). Rui Yan was sponsored by CCF-Tencent Open Research Fund.

\bibliographystyle{aaai}\bibliography{aaai}

\subsection{Appendix: Dimension Influence}
Table \ref{table_autoencoder} shows results of auto-encoder in paper-news task. With the increment of encoder (decoder) dimension, ability to recover source sequence increases. So content preservation is larger and transfer strength is smaller (more indeterminacy decreasing).

Table \ref{table_multi_decoder} shows results of multi-decoder in paper-news task. Similar to auto-encoder, with the dimension increasing, it shows higher ability to recover sentence.

Table \ref{table_style_embedding} shows results of style-embedding in paper-news task. In most cases, encoder dimension has little influence on results, However, style embedding dimension has a large influence on results. Larger style embedding dimension (also decoder dimension), more content preserved. The performance of decoder to recover sentence increases with dimension.

%table_autoencoder
\begin{table}[htb]
\centering
\begin{tabular}{|c|c|c|}
\hline
\makecell{encoder\\(decoder)\\dimension}   & \makecell{transfer\\strength}  & \makecell{content\\preservation}  \\ \hline
32  & 0.035   & 0.9118  \\
64  & 0.021   & 0.9524  \\
128 & 0.0135  & 0.9837  \\ \hline
\end{tabular}
\caption{Results for auto-encoder in paper-news task. Word embedding dimension is 64.}
\label{table_autoencoder}
\end{table}

%table_multi_decoder
\begin{table}[htb]
\centering
\begin{tabular}{|c|c|c|}
\hline
\makecell{encoder\\(decoder)\\dimension}   & \makecell{transfer\\strength}  & \makecell{content\\preservation}  \\ \hline
32  & 0.9915 & 0.8335 \\
64  & 0.9705 & 0.8440 \\
128 & 0.842  & 0.8705 \\ \hline
\end{tabular}
\caption{Results for multi-decoder in paper-news task. Word embedding dimension is 64.}
\label{table_multi_decoder}
\end{table}

%table_style_embedding
\begin{table}[htb]
\centering
\begin{tabular}{|c|c|c|c|}
\hline
\makecell{encoder\\dimension}   & \makecell{style\\embedding\\dimension} & \makecell{transfer\\strength}  & \makecell{content\\preservation}  \\ \hline
32  & 32  & 0.593   & 0.8966   \\
32  & 64  & 0.465   & 0.9157   \\
32  & 128 & 0.282   & 0.9435   \\
64  & 32  & 0.596   & 0.8957   \\
64  & 64  & 0.485   & 0.9153   \\
64  & 128 & 0.2745  & 0.9440   \\
128 & 32  & 0.595   & 0.8960   \\
128 & 64  & 0.4825  & 0.9146   \\
128 & 128 & 0.267   & 0.9439   \\ \hline
\end{tabular}
\caption{Results for style-embedding in paper-news task. Word dimension is 64. Decoder dimension equals to summing encoder dimension and style embedding dimension.}
\label{table_style_embedding}
\end{table}

\end{document}